\documentclass[runningheads]{llncs}
\usepackage{graphicx}
\usepackage{algorithm}
\usepackage[noend]{algpseudocode}
\usepackage{newtxmath}
\usepackage{comment}
\usepackage{circledsteps}
\usepackage{hyperref}
\usepackage{float}

\renewcommand{\algorithmicensure}{\textbf{Server executes:}}

\begin{document}
\title{Federated Neural Topic Models} 
%
%
\author{Lorena Calvo-Bartolomé \and
Jerónimo Arenas-García}


%
%
\institute{Universidad Carlos III de Madrid, Leganés, Spain\\
\email{\{lcalvo@pa,jarenas@ing\}uc3m.es}}
\maketitle              
\begin{abstract}
Over the last years, topic modeling has emerged as a powerful technique for organizing and summarizing big collections of documents or searching for particular patterns in them. However, privacy concerns may arise when cross-analyzing data from different sources. Federated topic modeling solves this issue by allowing multiple parties to jointly train a topic model without sharing their data. While several federated approximations of classical topic models do exist, no research has been conducted on their application for neural topic models.
To fill this gap, we propose and analyze a federated implementation based on state-of-the-art neural topic modeling implementations, showing its benefits when there is a diversity of topics across the nodes' documents and the need to build a joint model. In practice, our approach is equivalent to a centralized model training, but preserves the privacy of the nodes. Advantages of this federated scenario are illustrated by means of experiments using both synthetic and real data scenarios.

\end{abstract}

\section{Introduction}
\label{sec:intro}
Topic modeling refers to a common set of statistical tools to automatically extract the hidden structures of collections of unstructured text documents. Whereas Bayesian-based topic models (BTMs) led by Latent Dirichlet Allocation (LDA) \cite{blei2003latent} have been the main line of research in topic analysis for nearly two decades, today's new algorithmic developments mostly pertain to neural topic models (NTMs) \cite{miao2016neural,srivastava2017autoencoding}, aiming at improving scalability and, ideally, achieving higher-quality topics. More recently, there has been a significant focus on enhancing NTMs by incorporating pre-trained language models, which can capture more nuanced aspects of linguistic context through contextual word- and sentence-vector representations \cite{bianchi-etal-2021-pre,bianchi-etal-2021-cross,grootendorst2022bertopic,chaudhary2020topicbert}.

Many fields benefit from topic models, including Science, Technology, and Innovation (STI). Several studies (e.g., \cite{bonaccorsi2022exploring,nichols2014topic,talley2011database}) have shown that topic modeling can provide valuable insights into STI documents. By analyzing these documents' topics, researchers can answer many critical questions, such as comparing projects funded by different agencies, identifying research strengths of specific regions or organizations, and understanding the trends in a particular field.

For instance, topic modeling can help compare and contrast the research focus of different funding agencies and their contributions to a particular field. Additionally, it can aid in identifying specific regions or institutions' research strengths, such as the top research areas of a university or a country. This information is valuable for policymakers and funding agencies to make informed decisions about where to invest research resources.

However, obtaining high-quality topic models poses some challenges when we wish to build a shared topic model for comparison among several document collections, but privacy constraints apply. This limitation is frequently encountered in the STI analysis domain when processing collections of documents owned by different parties (e.g., funding agencies) that are either unwilling or do not have permission to share their documents without assurances of data privacy because of confidentiality, strategic reasons, or subjections to personal privacy owing to regulations such as the European Union General Data Protection Regulation (GDPR)\footnote{\url{https://gdpr-info.eu/}} or intellectual property rights (IPRs).

Federated Learning (FL) is an approach for collaboratively training machine learning models without sharing local data, with typically several clients coordinated with one or more central servers functioning as a mediator for the settings agreement, privacy assurance, and aggregation of the nodes' updates \cite{konevcny2016federated2,konevcny2016federated,mcmahan2017communication,shokri2015privacy}. Some researchers have studied FL's application to topic modeling due to its decentralized data leveraging and privacy-preserving properties. Some have focused on the design of LDA-alike or NMF-based federated frameworks \cite{jiang2019federated,wang2020federated,jiang2021industrial,si2022federated}, while others have opted for the proposal of federated general-purpose topic models \cite{shi2020federated}. However, to the best of our knowledge, no research has been carried out on the federated implementation of NTMs.

In this paper, we propose a novel method for the federated training of NTMs, namely gRPC-based federated neural topic models (\texttt{gFedNTM}). Concretely, we work under the following assumption: the global knowledge gained by $L$ nodes contributing to the construction of a federated NTM with no information loss is equivalent to that of the same nodes sharing all their information, but without being restricted to complete information disclosure. We claim that the predictive and inference capabilities attained by both centralized and federated scenarios outperform that of a single model building its own topic model in a non-collaborative scenario. In short, our contributions are:

\begin{itemize}
    \item Proposal of a generic framework for the federated training of NTMs, \texttt{gFedNTM}, with specific implementations for \textit{ProdLDA} \cite{srivastava2017autoencoding} and \textit{CombinedTM} \cite{bianchi-etal-2021-pre}. We provide our implementation in an open GitHub repository\footnote{\url{https://github.com/Nemesis1303/gFedNTM}}.
    \item Comparison of the knowledge gained in the centralized/federated scenarios with respect to the non-collaborative setup using synthetic data for objective evaluation.
    \item Qualitative evaluation in the context of STI analysis by utilizing different subsets from several Fields of Study of the Semantic Scholar dataset \cite{lo2019s2orc}.
\end{itemize}

This paper is a first proposal to train neural topic models in a privacy-preserving scenario. Further aspects relevant to federated learning, such as robustness to data loss or the presence of malicious nodes, will be addressed in future works.

The rest of the paper is organized as follows. In Section \ref{sec:background} we review some related work regarding NTMs and federated topic modeling. Section \ref{sec:methodology} extends these ideas and presents, to the best of our knowledge, the first implementation of NTMs with federated learning and satisfying privacy-preserving constraints. The implemented method has been released under a permissive license. Experimental work is carried out in Section \ref{sec:experiments}. Finally, Section \ref{sec:conclusions} presents the main conclusions of our work and some lines for future research.

\section{Background}
\label{sec:background}
\subsection{Neural Topic Models}
NTMs provide the $K$ main themes of a corpus of $D$ documents with vocabulary size $V$ in terms of the per topic-word $\boldsymbol{\beta}_{1:K}$ and the per document-topic $\boldsymbol{\theta}_{1:D}$ distributions. Yet, unlike BTMs, they estimate the latent distributions using neural networks rather than approximate variational Bayes or Gibbs' sampling procedures.

Product-of-Experts LDA (ProdLDA) \cite{srivastava2017autoencoding} consists of an encoder-decoder architecture with an inference network mapping the \textit{bag of words} (BoW) document representations into a continuous latent representation and a decoder network reconstructing this BoW. Its generative process is akin to LDA's \cite{blei2003latent}, but the Dirichlet prior is approximated via Gaussian distributions, and a weighted product-of-experts \cite{hinton2002training} replaces the multinomial distribution over individual words aiming at more convenient NN training, and higher quality topics that are better aligned with human judgment.

Contextualized Topic Models (CTMs) build over ProdLDA leveraging the integration of prior knowledge through contextualized embeddings. Among them, CombinedTM \cite{bianchi-etal-2021-pre} combines ProdLDA's input representation with SBERT embeddings \cite{reimers2019sentence}, and ZeroShotTM \cite{bianchi-etal-2021-cross} utilizes only the text embeddings. The inclusion of SBERT embeddings provides a lot of {\em a priori} information about language semantics beyond the actual document collection on which the topic model is trained. As a consequence, CTMs typically obtain significantly more coherent topics than their BTM counterparts.

\subsection{Federated Topic Modeling}

Previous works on federated topic modeling have followed a horizontal approach (i.e., parties holding different training samples but sharing the same feature space) to train models from the corpus owned by different parties. For example, \cite{jiang2019federated} proposed FTM, a Metropolis Hastings-based framework designed for the collective training of LDA-alike topic models; \cite{shi2020federated} introduced PC-TD, another federated topic model, but with a federated EM-based inference framework. More recently, \cite{si2022federated} presented FedNMF, an NMF-based topic modeling framework. As far as we can tell, whereas preceding algorithms have tailored BTM approaches by combining their traditional inference processes with secure aggregation protocols, there are no previous works building topic models in a federated setup using NTMs as the supporting technology.

One issue to overcome in an NTM-based federated topic modeling approach is model variability across the network, i.e., local nodes converging to different solutions. Two possible strategies would be: 1) to enforce convergence to a common solution through coordination with a server; and 2) to let each node learn its own solution and then learn a model that allows their mapping. Here, we opted for the first approach.

\section{Federated Neural Topic Models}
\label{sec:methodology}
\subsection{Problem statement}
\label{subsec:problem_statement}
Given a collection of $L$ local nodes (clients) and one central node (server), we consider the task of training a joint topic model that allows topic and document cross-comparison. The set of clients is denoted by $\mathcal{N}$. Each client $N_\ell$ ($\ell = 1, \dots, L$) holds a corpus $C_\ell$ with vocabulary $V_\ell$ in BoW format, i.e., a mapping between the words present in $C_\ell$ and the occurrence with which they appear. $V$ refers to the merge of all vocabularies, meaning that the set of all terms present in all vocabularies is kept (without repetitions) with weighted frequencies reflecting their overall presence across all nodes. The concatenation of all nodes' corpus is denoted as $C$.

We wish to learn a global NTM for corpus $C$ without clients sharing their local corpus $C_\ell$ among each other or with the server. We also wish to prove that such a global model, characterized by a network with weights $\mathbf{W}$, increases the knowledge that each client $N_\ell$ would acquire w.r.t. a non-collaborative topic model learned just from its own corpus $C_\ell$.

Therefore, our work in this paper considers three different scenarios depicted in Fig.~\ref{fig:all_scenarios}:

\begin{figure*}[!h]
    \centering
    \includegraphics[width=.85\textwidth]{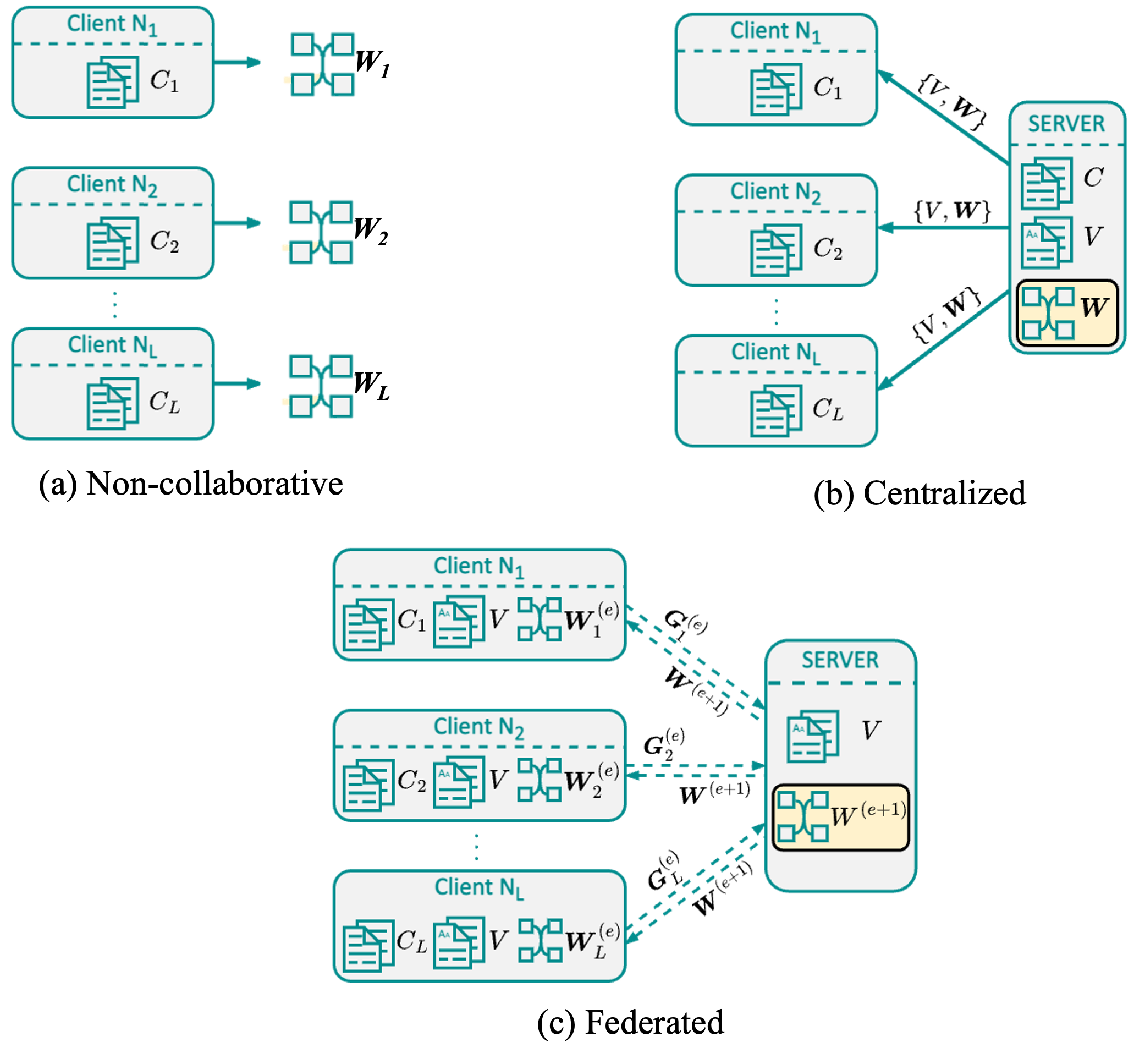}
    \caption{Graphical representation of the three  scenarios. (a) Non-collaborative: $L$ independent topic models are generated after training, one per node. (b) Centralized: The server trains a unique topic model using the information received from all nodes. Privacy is lost. (c) Federated: The nodes and the server collaboratively build a unique neural topic model using a process that preserves the privacy of the local corpora}.
    \label{fig:all_scenarios}
\end{figure*}

\begin{enumerate}
    \item \textbf{Non-collaborative scenario.} Each client $N_\ell$ builds its own topic model with its local corpus $C_\ell$. This scenario gives rise to $L$ different topic models, one per node.
    \item \textbf{Centralized scenario with a trusted server}. Clients share their corpus with the server, which learns a global model $\mathbf{W}$ from the concatenated corpus $C$. After the training is complete, the server sends the overall topic model and a common vocabulary, $\mathbf{W}$ and $V$, respectively, back to the clients. 
    \item \textbf{Centralized federated scenario with a non-trusted server}. Like the previous case, this scenario aims at building a common topic model from the concatenated corpus $C$, but the main difference is how the computation is split across the nodes, allowing the server to build such a global model without accessing the clients' corpora. Privacy-preserving is achieved at the cost of increased communication exchanges between the server and the local nodes.
\end{enumerate}

The third scenario described above is the one we consider in this paper. The next subsection describes our approach to Neural Topic Modeling for such a scenario.

\subsection{Federated Neural Topic Modeling (gFedNTM)}
We propose \texttt{gFedNTM}, a general federated framework for the privacy-preserving training of neural topic models. \texttt{gFedNTM} currently supports implementations for two state-of-the-art neural topic models, namely ProdLDA \cite{srivastava2017autoencoding} and CTM \cite{bianchi-etal-2021-pre,bianchi-etal-2021-cross}, but can be easily extended to other NTM-based algorithms. For the server-client communication, Google Remote Procedure Call (gRPC)\footnote{\url{https://grpc.io/}} is utilized. Our algorithm is available on GitHub under a permissive license\footnote{\label{note1}GitHub URL omitted to respect double-blind review.}.

\begin{figure*}[!h]
    \centering    \includegraphics[width=\textwidth]{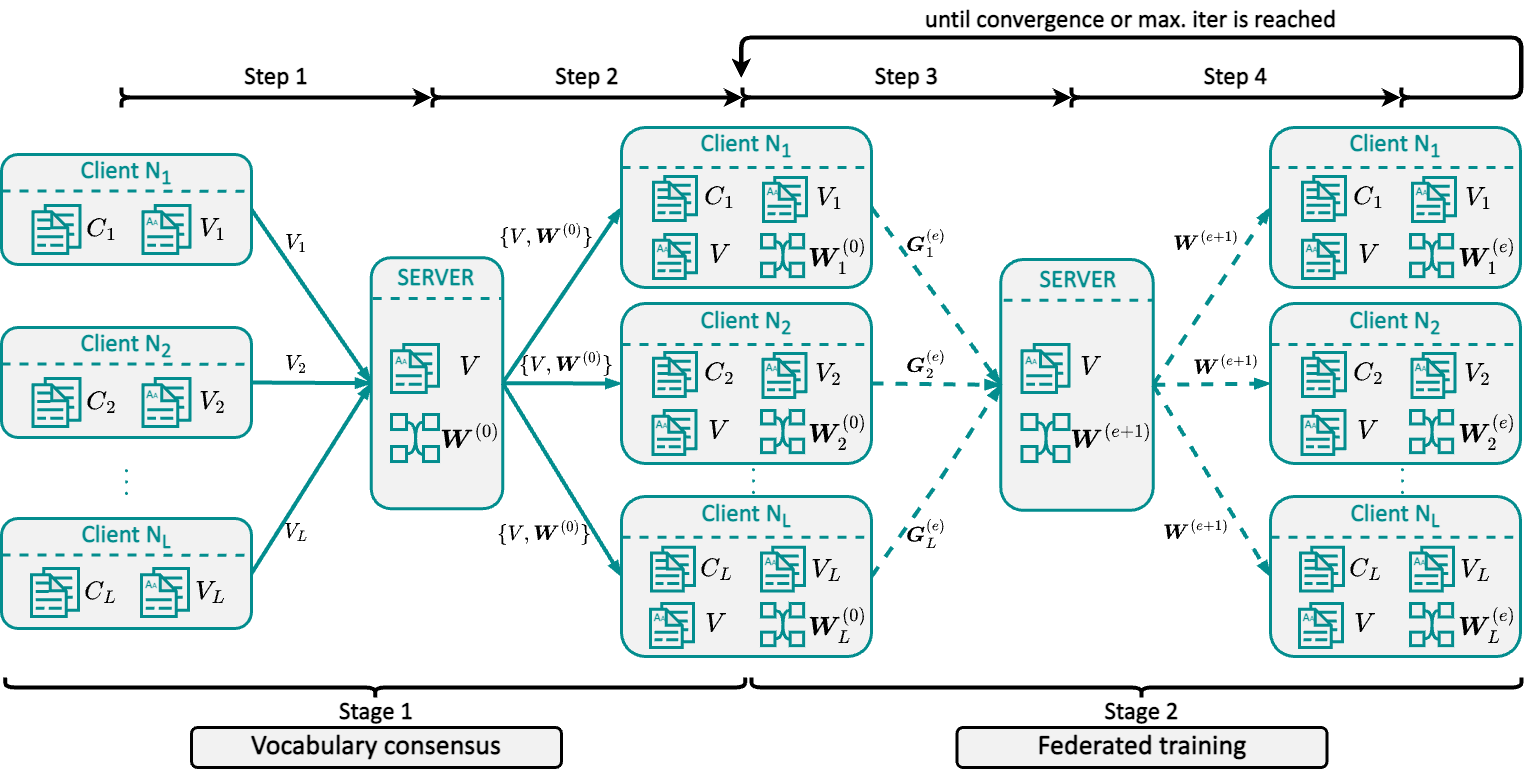}
    \caption{Training workflow under the federated scenario. The server only has access to the nodes' vocabularies $V_\ell$ and local gradients $\boldsymbol{G}_\ell^{(e)}$.}
    \label{fig:my_label}
\end{figure*}

Fig.~\ref{fig:my_label} illustrates the training process of the proposed federated topic model implementation. In our method, training takes place in two differentiated sequential stages: 1) \textit{vocabulary consensus}, and 2) \textit{federated training}.

In the \textit{vocabulary consensus stage} nodes compute and send their vocabularies to the server $\{V_\ell,\;\ell = 1, \dots, L\}$ (Step 1 in Fig.~\ref{fig:my_label}). Once the server receives the vocabularies of all nodes, it merges them into a common vocabulary $V$, which is used for initializing the global model with weights $\boldsymbol{W}^{(0)}$. Finally, the server sends both the common vocabulary and the initialized network weights back to the clients (Step 2 in Fig.~\ref{fig:my_label}).

During the \textit{federated training stage}, each client trains its local model using its own data and sends the model updates to the server for aggregation. The server integrates these updates to obtain updated global model parameters, which are sent back to the clients for further local training. This process is iteratively repeated until a stopping criterion is met, which is either when the relative variation of the weights in the networks no longer improves or when a predefined maximum number of iterations is reached.

For the training itself, we utilize Synchronous Stochastic Optimization (Sync-Opt) \cite{chen2016revisiting} (Steps 3 and 4 in Fig.~\ref{fig:my_label}), which works by synchronizing the model parameters across all the participating clients after each mini-batch update. At each mini-batch step, the server waits for all the clients to send their gradients $\boldsymbol{G}_{\ell}^{(e)}$ and combines them to obtain a global gradient $\boldsymbol{G}^{(e)}$ via an aggregation function $\boldsymbol{Agg}(\cdot)$,

\begin{equation}
    \boldsymbol{G}^{(e)} \leftarrow \boldsymbol{Agg}(\{\boldsymbol{G}_\ell^{(e)}, \ell=1,...,L\}).
\end{equation}

Although there are several options for aggregating the gradients, we use the weighted average function in our implementation. This means that the gradients contributed by each client are averaged proportionally to the number of training samples (i.e., documents) in each client's mini-batch, i.e.,

\begin{equation}
    \boldsymbol{G}^{(e)} = \frac{\sum_{\ell =1}^{L} n_\ell \cdot  \boldsymbol{G}_\ell^{(e)}}{\sum_{\ell =1}^{L} n_\ell},
\end{equation}
where $n_\ell, ~\ell=1,\dots, L$ is the number of samples in each client $C_{\ell}$'s mini-batch.

This function ensures that clients with larger mini-batches contribute more to the global gradient, which helps to balance the contribution of each client to the training process.

After aggregation takes place, the global gradient $\boldsymbol{G}^{(e)}$ is then utilized, also in the server side, to update the current global model parameters $\boldsymbol{W}^{(e)}$ using a standard gradient update with step size parameter $\lambda$: 
\begin{equation}
    \boldsymbol{W}^{(e+1)} \leftarrow \boldsymbol{W}^{(e)} - \lambda \cdot \boldsymbol{G}^{(e)}.
\end{equation}

The updated model parameters are then sent to all clients.

This approach allows for efficient and synchronized training across all clients, with the effective batch size being the sum of all the clients' mini-batch sizes. The entire process is described in Algorithm \ref{algorithm:gfedNTM}, which illustrates the operation of both the server and the client nodes. Overall, \texttt{gFedNTM} enables multiple clients to jointly train a global topic model without compromising the privacy of their local data. This framework is particularly useful for scenarios where the data is distributed across multiple clients and cannot be shared due to privacy concerns. 

\begin{algorithm}[!h]
     \footnotesize
     \begin{algorithmic}[1]
        \Require $\mathcal{C},K$
        \Ensure
        \State{\textit{// stage \Circled{1}: vocabulary consensus}}
        \For{each client $N_\ell \in \mathcal{N}$ \textbf{in parallel}} 
            \State{$V_\ell \leftarrow$} \Call{getClientVocab}{$N_\ell$}
        \EndFor
        \State{$V \leftarrow merge(\{V_\ell\}) $}
        \State{Send $V$ to each $N_\ell \in \mathcal{N}$}
        \State{Initialize $\boldsymbol{W}^{(0)}$}
        \State{\textit{// stage \Circled{2}: federated training}} 
        \While{training: $e=0,...,I$}
            \For{each client $N_\ell \in \mathcal{N}$ \textbf{in parallel}}
                \State{$\boldsymbol{G}_\ell^{(e)} \leftarrow$ \Call{getClientGrad}{$N_\ell,\boldsymbol{W}^{(e)}$}}
            \EndFor
            \State{$\boldsymbol{G}^{(e)} \hspace{4mm} \leftarrow \boldsymbol{Agg}(\{\boldsymbol{G}_\ell^{(e)}, \ell=1,...,L\})$}
            \State{$\boldsymbol{W}^{(e+1)} \leftarrow \boldsymbol{W}^{(e)} - \lambda \cdot \boldsymbol{G}^{(e)}$}
        \EndWhile
        
        \renewcommand{\algorithmicensure}{\textbf{Client $N_\ell$ executes:}}
        \Ensure \setcounter{ALG@line}{0}
        \Function{getClientVocab}{$N_\ell$} 
            \State{return $V_\ell$ to server}
        \EndFunction

        \Function{getClientGrad}{$N_\ell, \boldsymbol{W}^{(e)}$}
            \State{Select mini-batch $b$}
            \State{$\boldsymbol{W}_\ell^{(e)} \leftarrow \boldsymbol{W}^{(e)}$}
            \State{$\boldsymbol{G}_\ell^{(e)} \leftarrow 
                    \nabla L(\boldsymbol{W}_{l}^{(e)};b)$}
            \State{return $\boldsymbol{G}_\ell^{(e)}$ to server}
        \EndFunction
    \end{algorithmic}
    
    \caption{\textbf{gFedNTM.} $I$ is the maximum number of iterations, and $\lambda$ is the learning rate.}
    \label{algorithm:gfedNTM}
\end{algorithm}

\section{Experiments}
\label{sec:experiments}
In this section, we provide results on synthetic and real datasets to check the validity of our proposal. We use the synthetic data to illustrate the advantages of centralized/federated model training with respect to the non-collaborative scenario since, in this case, having a ground truth enables objective evaluation. Analysis of the real datasets offers a more challenging situation where the evaluation will be pursued using Word Mover's Distance (WMD) across topics \cite{Pele2009FastAR}.

\subsection{Collaborative \textit{vs} non-collaborative topic modeling}
\label{subsec:centr}
In this first subsection, we focus on comparing the models learned under collaborative and non-collaborative setups. For the collaborative scenario, we use Scenario 2 described in Subsection \ref{subsec:problem_statement}, i.e., a centralized implementation. Yet, we have checked that \texttt{gFedNTM} achieves the same results (as theoretically expected). 

\subsubsection{Experimental setup and evaluation criteria}
For this first set of experiments, we use synthetic documents, generated according to the generative model inherent to LDA \cite{blei2003latent}. This allows an objective evaluation of the models by comparing the topics and document representations obtained after training with the true generative model. We consider a network with $L=5$ nodes and a common vocabulary of $5\,000$ artificial terms (i.e., with no semantic meaning, such as \textit{term56}, \textit{term125}, etc.). The total number of topics is set at $K=50$. To simulate topic diversity and coincidence across nodes, we consider that $K^{'}$ topics are shared by all nodes, whereas $(K-K')/L$ topics are private for each node. Each node generates a total of $10\,000$ training and $1\,000$ validation documents using the true underlying generative model of LDA, each document with random length drawn uniformly in the range $[150,250]$. Hyperparameters of the model were adjusted as follows: Dirichlet hyperparameter for the document representations was fixed to $\alpha = 50/K$ whereas different values were used for the hyperparameter of the Dirichlet for the generation of topic distributions ($\eta$).

We performed two sets of experiments: 
\begin{enumerate}
    \item[A)] we varied $K^{'}\in \{5,10,15,30,40\}$ and fixed the topic prior $\eta$ to $0.01$;
    \item[B)] we set $K^{'}=10$ and varied $\eta\in\{0.01, 0.02, 0.03, 0.04, 0.08, 1\}$.
\end{enumerate}
With this, we seek to quantitatively evaluate the performance gain of the centralized scenario as a function of the number of common topics ($K^{'}$) shared across nodes (experimental setting A) and of the specificity of the topics' vocabularies (higher for smaller values of $\eta$ in setting B).

For each setting, we trained a local model at each node, as well as a centralized model at the server with the $50\,000$ documents (i.e., the concatenation of all the nodes' documents). Then we inferred the topic representation of the $5\,000$ validation documents in both scenarios. Regarding the ProdLDA's parameters, we utilized the default settings provided by the authors \cite{srivastava2017autoencoding}. For the early stopping of the neural network, we split the data into a $75:25$ ratio.

Since in this scenario we know the true per topic and per document distribution vectors ($\boldsymbol{\beta}_{1:K}$ and $\boldsymbol{\theta}_{1:D}$), objective evaluation can be based on the comparison of the latent variables obtained after ProdLDA training (using either centralized or non-collaborative scenarios) and the known true values given by the generative process. Two quantitative assessment criteria are proposed for this:

\begin{enumerate}
    \item \textbf{Document similarity-based score (DSS).} Let the similarity between any two documents be computed from their Hellinger distance, as:
    \begin{equation}
        w_{ij} = 1 - H^2 (\boldsymbol{\theta}_i, \boldsymbol{\theta}_j)= \sqrt{\boldsymbol{\theta}_i}^T\sqrt{\boldsymbol{\theta}_j}.
    \end{equation}
    Then, the document similarity-based score of a given topic model is proposed as the sum of all differences between the similarities of every pair of documents and the corresponding true similarities calculated from the ground truth document representations, divided by the number of documents, i.e.,
    \begin{equation}
        \text{DSS} = \frac{1}{D} \sum_{i=1}^{D} \sum_{j=1, j\neq i}^{D} \mid w_{ij}^{(\text{true})} - w_{ij}^{(\text{inferred})} \mid,
    \end{equation}
    where $D$ is the total number of documents. The smaller the DSS, the better model in relation to document characterization.
    \item \textbf{Topic similarity score (TSS).} Likewise, we can calculate the similarity between any pair of topics using the same similarity function defined above. Then, the topic similarity score between the true model and any learned model is calculated as the sum of similarities between every topic from the true model and the closest topic from the inferred model, i.e.,
    \begin{equation}
        \text{TSS} = \sum_{k=1}^K \max_{k'} \left[ 1 - H^2(\boldsymbol{\beta}_k^{(\text{true})}, \boldsymbol{\beta}_{k'}^{(\text{inferred})}) \right].
    \end{equation}
    The closer this indicator is to $K$, the better the alignment between the true and inferred models.  
\end{enumerate}

\begin{figure}[!h]
    \centering
    \includegraphics[width=\textwidth]{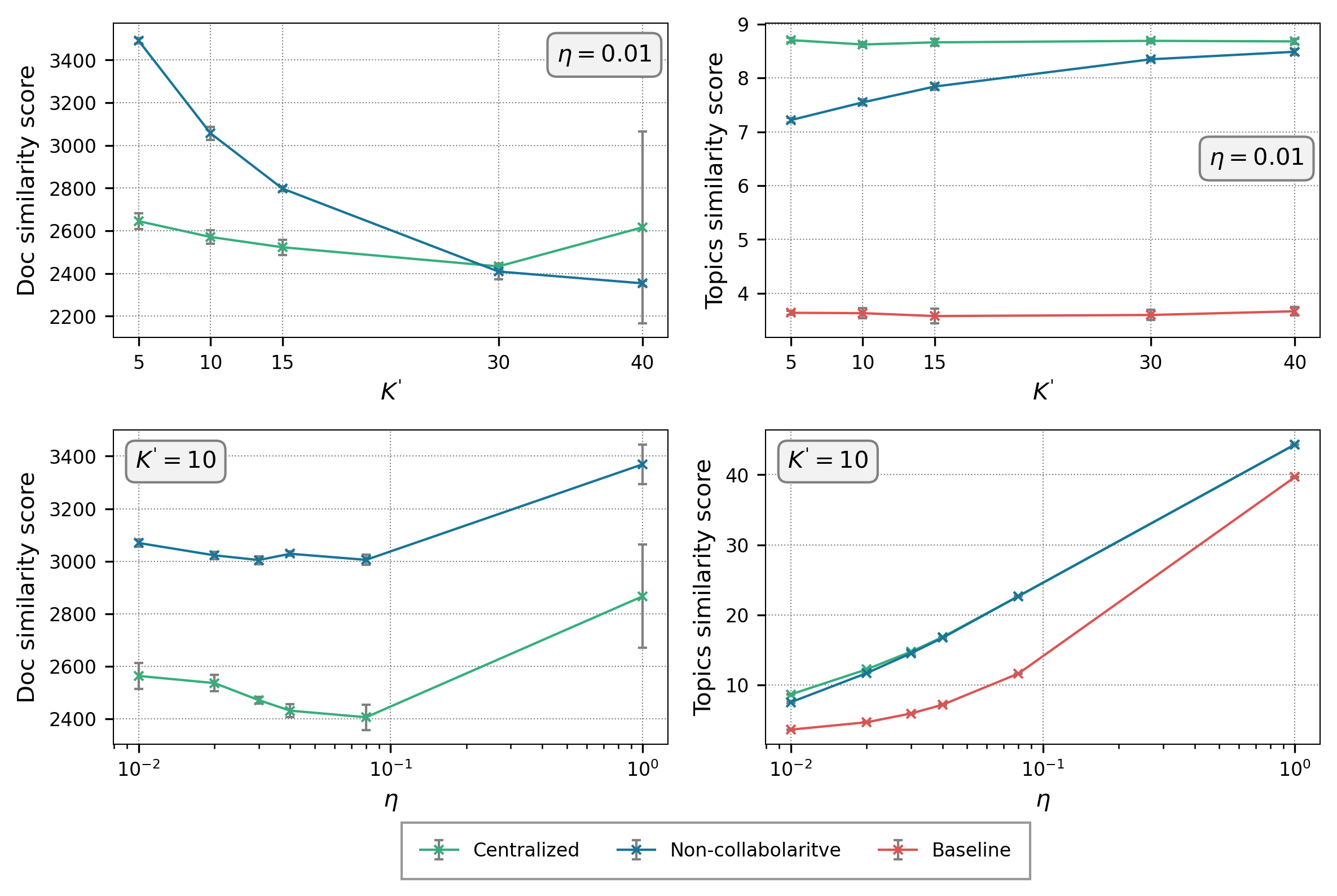}
    \caption{Document similarity-based (left) and topic similarity (right) scores for non-collaborative and centralized scenarios.}%
    \label{fig:sim_vs_nr_frozen}
\end{figure}

\subsubsection{Evaluation of results}
\label{subsec:eval_syn}
Fig.~\ref{fig:sim_vs_nr_frozen} displays the performance of the centralized and non-collaborative scenarios as a function of the number of shared topics $K'$ (experimental setting A, top row) and the Dirichlet hyperparameter $\eta$ (setting B, bottom row). All results have been averaged over 10 runs. For the TSS evaluation, we have also included a baseline for comparison. This baseline model is taken from the {\em a priori} distributions used to generate the true model. In other words, it provides the expected TSS between two independent synthetic models sampled from the same distribution, and provides the minimum TSS that should be achieved by any informed model.


Analyzing the results for varying numbers of common topics (top row of Fig. \ref{fig:sim_vs_nr_frozen}), we can conclude that the centralized scenario has better inference capabilities than the non-collaborative scenario. As expected, this performance gain is more significant as the number of common topics decreases (i.e., smaller $K'$), since, in this situation, the local nodes' documents become more different. For larger $K'$ the nodes share many common topics, reducing the margin for a gain in performance; still, some advantages are visible in terms of TSS whereas the situation seems to reverse for the DSS and $K'=40$. However, notice that in this situation we observe a large variability of the DSS, making results inconclusive. This larger variability is probably due to the fact that nodes have a very small number of topics of their own, making the runs more different.

Though the centralized scheme still shows some advantage for varying $\eta$ (see bottom row of Fig. \ref{fig:sim_vs_nr_frozen}), the performance gain is somewhat less significant, especially in terms of the topic similarity score. This is primarily due to ProdLDA rather than the scenario itself since the per-topic distribution vectors inferred by ProdLDA tend to have many active words per topic, thus reducing the margin for performance gains.


These findings support our motivation for proposing algorithms that support the need for building a consensus topic model that can be shared across all client nodes. In the next subsection, we will reinforce this conclusion with experiments carried out on real documents.

\subsection{Federated topic modeling on real data}

In this second subsection, we compare the models learned on real data under federated {\tt gFedNTM} and non-collaborative setups.

\subsubsection{Experimental setup and evaluation criteria}
We utilize five subsets from the Semantic Scholar Open Research Corpus \cite{lo2019s2orc} in the disciplines of Computer Science, Economics, Sociology, Philosophy, and Political Science, assigning each category's documents (the concatenation of the articles' textual fields) to a different client. After preprocessing, we retain $732\,039$, $616\,261$, $440\,139$, $133\,545$ and $304\,195$ documents. We simulated a federated scenario through {\tt gFedNTM} with CTM as the underlying algorithm and trained two federated models with 10 and 25 topics. The maximum number of federated training iterations was 100, and CTM's parameters were set to the default values suggested by the authors \cite{bianchi-etal-2021-pre}. Additionally, we trained non-collaboratively a CTM topic model at each node on its corpus.

The intention here is to qualitatively evaluate the ability of \texttt{gFedNTM} to build a common topic model mapping all the clients' collections. Working with real data, we exploit semantic topic similarity leaning on the Word Mover’s Distance (WMD)\footnote{\url{https://radimrehurek.com/gensim/auto_examples/tutorials/run_wmd.html}} \cite{kusner2015word}. We specifically assess how effectively all models (federated and node-specific) capture the specific topics of the collections at each node by defining the Average Minimum WMD (AMWMD) between the topics of collection $C_\ell$ (that result from the non-collaborative topic model at $N_\ell$) and the topics of any other model as
\begin{equation}
    \text{AMWMD}^{(l, eval)} = \sum_{k=1}^K \min_{k'} \text{WMD}(\text{TD}_k^{(l)}, \text{TD}_{k'}^{(eval)}),
\end{equation}
where $\text{TD}_k^{(l)}$ is the text description of topic $k$ from the model at node $N_\ell$ and $\text{TD}_{k'}^{(eval)}$ that of topic $k'$ from the model being evaluated, either a node-specific model or a federated model. Obviously, $\text{AMWMD}^{(l, eval)} = 0$ when the model being evaluated is the specific model for node $l$. However, we expect to see that federated models tend to work better on average and cover more accurately the topics of all 5 nodes.

\subsubsection{Evaluation of results}
\begin{figure}[!ht]
    \centering
    \includegraphics[width=\textwidth]{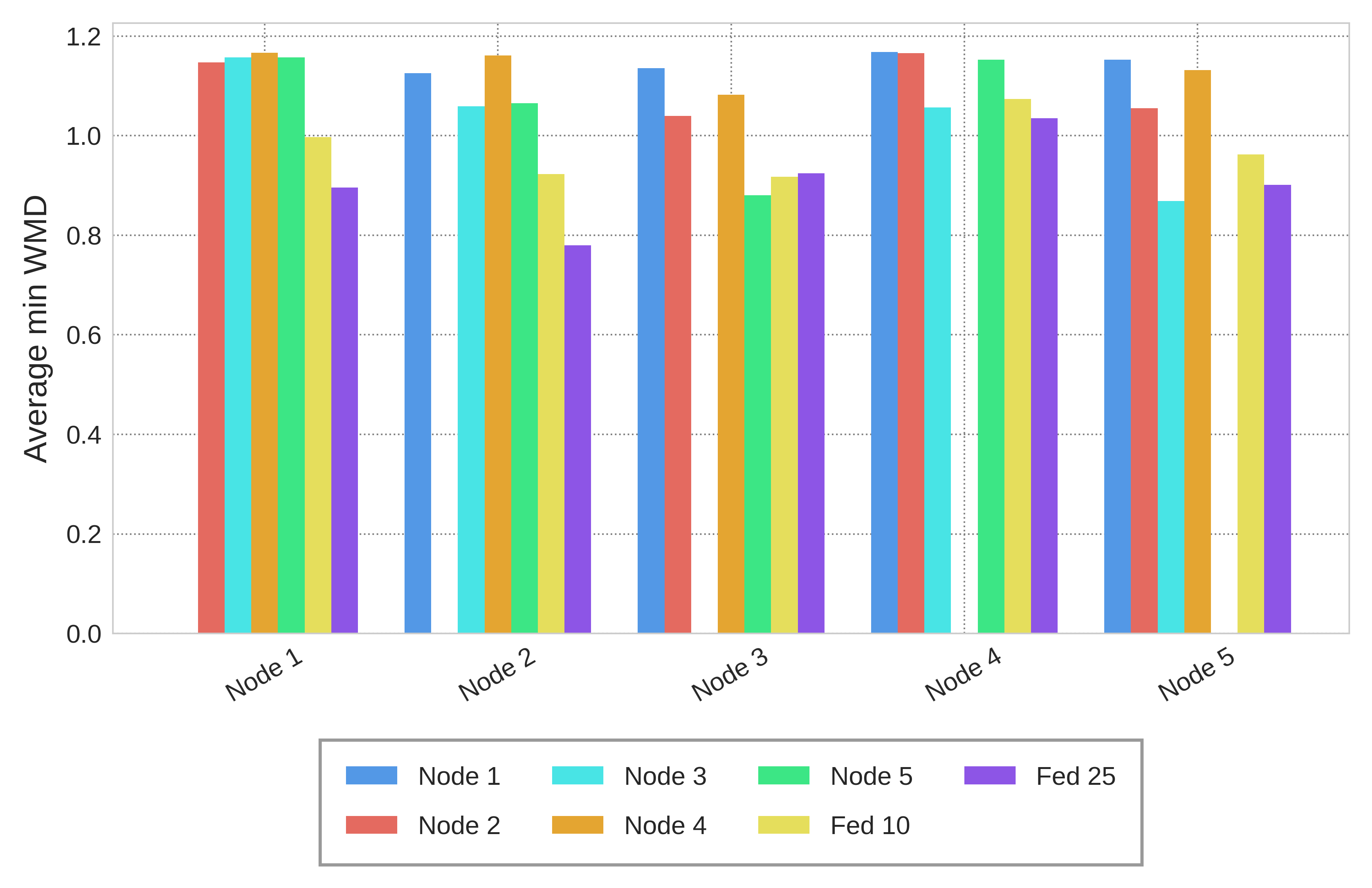}
    \caption{AMWMD between the nodes' and federated models' topics.}
    \label{fig:barplot}
\end{figure}

Fig.~\ref{fig:barplot} displays the AMWMD between the topics of the non-collaborative at each node and the federated models. We can observe how the 10-topics federated model already shows an improvement in characterizing the nodes' topics than any other individual node's model, which is further improved with the 25-topics federated model. This knowledge gain reinforces the conclusion attained in Subsection \ref{subsec:centr} of the viability of constructing consensus topic models, and we can conclude that {\tt gFedNTM} allows appropriate learning of the topics belonging to each client without the need for document sharing among them or with the server. 

\section{Conclusions and Future work}
\label{sec:conclusions}
In this paper, we have presented a novel framework for the federated training of NTMs when building topic models for cross-comparison is required. We have justified this approach by exposing the information gain attained with a centralized scenario w.r.t. a non-collaborative on synthetic data and then reinforced this gain through experiments on real data with {\tt gFedNTM}. Future work should focus on the improvement of {\tt gFedNTM} to make it robust to missing data and network failures, as the present work assumed perfect communication among network nodes. Further, it will be extended to consider a decentralized federated scenario, i.e., the clients coordinate themselves to obtain the global model, without an intermediate server.

%
%
%
%
\bibliographystyle{splncs04}
\bibliography{refs}

\begin{thebibliography}{10}
\providecommand{\url}[1]{\texttt{#1}}
\providecommand{\urlprefix}{URL }
\providecommand{\doi}[1]{https://doi.org/#1}

\bibitem{bianchi-etal-2021-pre}
Bianchi, F., Terragni, S., Hovy, D.: Pre-training is a hot topic:
  Contextualized document embeddings improve topic coherence. In: Proc. 59th
  Annual Meeting of the Association for Computational Linguistics and the 11th
  Intl. Joint Conf. Natural Language Processing (Volume 2: Short Papers). pp.
  759--766 (Aug 2021)

\bibitem{bianchi-etal-2021-cross}
Bianchi, F., Terragni, S., Hovy, D., Nozza, D., Fersini, E.: Cross-lingual
  contextualized topic models with zero-shot learning. In: Proc. 16th Conf. of
  the European Chapter of the Association for Computational Linguistics: Main
  Volume. pp. 1676--1683. Online (Apr 2021)

\bibitem{blei2003latent}
Blei, D.M., Ng, A.Y., Jordan, M.I.: Latent dirichlet allocation. Journal of
  machine Learning research  \textbf{3}(Jan),  993--1022 (2003)

\bibitem{bonaccorsi2022exploring}
Bonaccorsi, A., Melluso, N., Massucci, F.A.: Exploring the antecedents of
  interdisciplinarity at the european research council: a topic modeling
  approach. Scientometrics pp. 1--31 (2022)

\bibitem{chaudhary2020topicbert}
Chaudhary, Y., Gupta, P., Saxena, K., Kulkarni, V., Runkler, T., Sch{\"u}tze,
  H.: {TopicBERT} for energy efficient document classification. arXiv preprint
  arXiv:2010.16407  (2020)

\bibitem{chen2016revisiting}
Chen, J., Pan, X., Monga, R., Bengio, S., Jozefowicz, R.: Revisiting
  distributed synchronous {SGD}. arXiv preprint arXiv:1604.00981  (2016)

\bibitem{grootendorst2022bertopic}
Grootendorst, M.: {BERTopic}: Neural topic modeling with a class-based {TF-IDF}
  procedure. arXiv preprint arXiv:2203.05794  (2022)

\bibitem{hinton2002training}
Hinton, G.E.: Training products of experts by minimizing contrastive
  divergence. Neural computation  \textbf{14}(8),  1771--1800 (2002)

\bibitem{jiang2019federated}
Jiang, D., Song, Y., Tong, Y., Wu, X., Zhao, W., Xu, Q., Yang, Q.: Federated
  topic modeling. In: Proc. 28th ACM Intl. Conf. information and knowledge
  management. pp. 1071--1080 (2019)

\bibitem{jiang2021industrial}
Jiang, D., Tong, Y., Song, Y., Wu, X., Zhao, W., Peng, J., Lian, R., Xu, Q.,
  Yang, Q.: Industrial federated topic modeling. ACM Transactions on
  Intelligent Systems and Technology (TIST)  \textbf{12}(1),  1--22 (2021)

\bibitem{konevcny2016federated2}
Kone{\v{c}}n{\`y}, J., McMahan, H.B., Ramage, D., Richt{\'a}rik, P.: Federated
  optimization: Distributed machine learning for on-device intelligence. arXiv
  preprint arXiv:1610.02527  (2016)

\bibitem{konevcny2016federated}
Kone{\v{c}}n{\`y}, J., McMahan, H.B., Yu, F.X., Richt{\'a}rik, P., Suresh,
  A.T., Bacon, D.: Federated learning: Strategies for improving communication
  efficiency. arXiv preprint arXiv:1610.05492  (2016)

\bibitem{kusner2015word}
Kusner, M., Sun, Y., Kolkin, N., Weinberger, K.: From word embeddings to
  document distances. In: International conference on machine learning. pp.
  957--966. PMLR (2015)

\bibitem{lo2019s2orc}
Lo, K., Wang, L.L., Neumann, M., Kinney, R., Weld, D.S.: {S2ORC}: The semantic
  scholar open research corpus. arXiv preprint arXiv:1911.02782  (2019)

\bibitem{mcmahan2017communication}
McMahan, B., Moore, E., Ramage, D., Hampson, S., Arcas, B.A.:
  Communication-efficient learning of deep networks from decentralized data.
  In: Artificial intelligence and statistics. pp. 1273--1282. PMLR (2017)

\bibitem{miao2016neural}
Miao, Y., Yu, L., Blunsom, P.: Neural variational inference for text
  processing. In: International conference on machine learning. pp. 1727--1736.
  PMLR (2016)

\bibitem{nichols2014topic}
Nichols, L.G.: A topic model approach to measuring interdisciplinarity at the
  national science foundation. Scientometrics  \textbf{100}(3),  741--754
  (2014)

\bibitem{Pele2009FastAR}
Pele, O., Werman, M.: Fast and robust earth mover's distances. 2009 IEEE 12th
  International Conference on Computer Vision pp. 460--467 (2009)

\bibitem{reimers2019sentence}
Reimers, N., Gurevych, I.: Sentence-bert: Sentence embeddings using siamese
  bert-networks. arXiv preprint arXiv:1908.10084  (2019)

\bibitem{shi2020federated}
Shi, Y., Tong, Y., Su, Z., Jiang, D., Zhou, Z., Zhang, W.: Federated topic
  discovery: A semantic consistent approach. IEEE Intelligent Systems
  \textbf{36}(5),  96--103 (2020)

\bibitem{shokri2015privacy}
Shokri, R., Shmatikov, V.: Privacy-preserving deep learning. In: Proceedings of
  the 22nd ACM SIGSAC conference on computer and communications security. pp.
  1310--1321 (2015)

\bibitem{si2022federated}
Si, S., Wang, J., Zhang, R., Su, Q., Xiao, J.: Federated non-negative matrix
  factorization for short texts topic modeling with mutual information. arXiv
  preprint arXiv:2205.13300  (2022)

\bibitem{srivastava2017autoencoding}
Srivastava, A., Sutton, C.: Autoencoding variational inference for topic
  models. arXiv preprint arXiv:1703.01488  (2017)

\bibitem{talley2011database}
Talley, E.M., Newman, D., Mimno, D., Herr, B.W., Wallach, H.M., Burns, G.A.,
  Leenders, A., McCallum, A.: Database of {NIH} grants using machine-learned
  categories and graphical clustering. Nature Methods  \textbf{8}(6),  443--444
  (2011)

\bibitem{wang2020federated}
Wang, Y., Tong, Y., Shi, D.: Federated latent dirichlet allocation: A local
  differential privacy based framework. In: Proceedings of the AAAI Conference
  on Artificial Intelligence. vol.~34, pp. 6283--6290 (2020)

\end{thebibliography}

\end{document}